%% file: main.tex
\documentclass[runningheads]{llncs}

\usepackage[T1]{fontenc}
\usepackage{graphicx}
\usepackage{hyperref}
\usepackage{color}

\usepackage{tikz}
\usetikzlibrary{shapes,arrows}
\usepackage{graphicx}
\usepackage[section]{placeins}
\usepackage{booktabs}
\usepackage{makecell}
\usepackage{tabularx}
\usepackage{comment}
\usepackage{pgf}
\usepackage{subcaption}
\usepackage{comment}
\usepackage{xcolor}

\usepackage[ruled,vlined]{algorithm2e}

\usepackage{mwe}
\usepackage{tablefootnote}

\usepackage{algpseudocode}
\usepackage{booktabs}
\usepackage{multirow}
\usepackage{eurosym}
\usepackage{amssymb}
\usepackage{amsmath}
\usepackage{wasysym}
\usepackage{mathtools}
\usepackage{siunitx,etoolbox}
\usepackage{listings}
\usepackage{float} 
\usepackage{tabularx}
\usepackage{xcolor}
\definecolor{mydarkgreen}{rgb}{0.0, 0.5, 0.0}

\usepackage{ragged2e}
\usepackage{array}
\usepackage{anyfontsize}
\usepackage{framed}
\usepackage[none]{hyphenat}
\usepackage{hyperref}
\usepackage{pgf}
\usepackage{enumitem} 
\usepackage{listings}
\usepackage{xcolor}
\usepackage{listings}

\usepackage[most]{tcolorbox}   
\usepackage{enumitem}          
\usepackage{listings}          
\usepackage{colortbl}  

\usepackage{amssymb}
\allowdisplaybreaks

\usepackage{booktabs}
\begin{document}
\title{Algorithmic Prompt-Augmentation for Efficient LLM-Based Heuristic Design for A* Search}
\titlerunning{Algorithmic Prompt-Augmentation for Efficient LLM-Based Heuristic Design}
%
\author{Thomas Bömer\inst{1}\orcidID{0000-0003-4979-7455} \and Nico Koltermann\inst{2}\orcidID{0009-0008-0359-0452} \and
Max Disselnmeyer\inst{1}\orcidID{0009-0008-5689-2235}\and
Bastian Amberg\inst{1}\orcidID{0000-0001-6715-3819}\and
Anne Meyer \inst{1}\orcidID{0000-0001-6380-1348}
}

\authorrunning{T. Bömer et al.}
%
\institute{
Karlsruhe Institute of Technology, Kriegsstraße 77, 76133 Karlsruhe, Germany 
\email{\{thomas.boemer,
max.disselnmeyer,
bastian.amberg,
anne.meyer\}}@kit.edu
\and
TU Dortmund University, Leonhard-Euler-Stra{\ss}e 5, 44227 Dortmund, Germany 
\email{\{nico.koltermann\}@tu-dortmund.de} 
}
\maketitle              
\begin{abstract}
\input{chapter/0_abstract}

\end{abstract}
\input{chapter/1_text}

\begin{credits}
\subsubsection{\ackname} This research was funded by the Ministry of Science, Research and Arts of the Federal State of Baden-Württemberg - InnovationCampus Future Mobility - funding code BUP59 FitLLM and the European Union - NextGenerationEU - funding code 13IK032I.

\end{credits}
%
%
%
\bibliographystyle{splncs04}
\bibliography{sources.bib}

\newpage

\end{document}

%% file: chapter/0_abstract.tex
Heuristic functions are essential to the performance of tree search algorithms such as A*, where their accuracy and efficiency directly impact search outcomes. Traditionally, such heuristics are handcrafted, requiring significant expertise. Recent advances in large language models (LLMs) and evolutionary frameworks have opened the door to automating heuristic design. In this paper, we extend the Evolution of Heuristics (EoH) framework to investigate the automated generation of guiding heuristics for A* search. 
We introduce a novel domain-agnostic prompt augmentation strategy that includes the A* code into the prompt to leverage in-context learning, named Algorithmic - Contextual EoH (A-CEoH).
To evaluate the effectiveness of A-CeoH, we study two problem domains: the Unit-Load Pre-Marshalling Problem (UPMP), a niche problem from warehouse logistics, and the classical sliding puzzle problem (SPP).
Our computational experiments show that A-CEoH can significantly improve the quality of the generated heuristics and even outperform expert-designed heuristics.

%% file: chapter/1_text.tex
\section{Introduction}
\label{sec:Introduction}
Combinatorial optimization problems play a crucial role in logistics and manufacturing. Solving such problems depends on heuristics to find high-quality solutions in a reasonable time. The design of such heuristics, however, is traditionally a knowledge-intensive and manual process, requiring deep domain expertise and extensive experimentation \cite{burke2013hyper}. This has motivated research into methods that can automatically generate heuristics.
\par
Recent advances in large language models (LLMs) have opened a promising new direction for automated heuristic design. Several frameworks have demonstrated how LLMs can serve as heuristic designers within an evolutionary loop \cite{romera2024mathematical,liu2024evolutionEoH,ye2024reevo}. In these frameworks, LLMs generate candidate heuristic functions, which are then evaluated on problem instances, and high-performing heuristics are iteratively mutated and refined. These methods have achieved encouraging results for constructive greedy search approaches. Especially, the design of constructive heuristics for the Traveling Salesperson Problem (TSP) and Online Bin Packing Problem (oBPP) seems to establish as a standard benchmark introduced by the first works \cite{romera2024mathematical,liu2024evolutionEoH} and adapted by various extensions and variations \cite{ye2024reevo,dat2024hsevo,2024efficientHERCULES} in this emerging research field.
\par
Current works in LLM-empowered automated heuristic design predominantly study constructive heuristics in optimization domains that always yield feasible solutions (even if far from optimal) like the TSP and oBPP. In contrast, many optimization domains impose hard constraints or state-dependent restrictions that limit feasible moves during search. In these cases, early decisions can strongly influence the reachable parts of the search space, making simple greedy approaches ineffective. A simple constructive heuristic cannot reliably find a solution \cite{bomer2025leveraging} or generalizes poorly \cite{sim2025beyond}. For such problems, more structured search algorithms such as the \texttt{A*} algorithm are required to efficiently explore the solution space while potentially maintaining optimality guarantees through an admissible heuristic \cite{hartAstar}.
\par
In this study, we investigate the potential of LLMs to automatically design heuristics suitable for guiding \texttt{A*} search. To explore this question, we focus on two problem domains with contrasting characteristics: the \textit{Unit-Load Pre-Marshalling Problem (UPMP)} and the \textit{Sliding Puzzle Problem (SPP)}.
The UPMP involves determining a sequence of moves to reorder unit loads in a block-stacking warehouse according to priority classes \cite{pfrommer2023solving,pfrommerboemer2024sorting,bomer2024sorting}. As a relatively new problem introduced in 2023, the UPMP is scarcely represented in LLM training data, and only a few handcrafted heuristics are currently available.
The SPP requires finding a sequence of moves that transforms a given initial configuration into a target goal state \cite{KORF198597}. In contrast to the UPMP, this problem is a well-established benchmark in algorithmic research, studied extensively since 1980s \cite{KORF198597}, with numerous heuristics proposed in both academic literature \cite{KORF20029,HANSSON1992207,hasan2023fifteen} and informal implementations on platforms such as GitHub.
Together, these two domains enable a balanced evaluation, covering both a practical, domain-specific problem with limited prior knowledge (UPMP) and a canonical, well-studied benchmark (SPP).
\par
While designing heuristics for problems such as the TSP and oBPP is useful for initially benchmarking new automated heuristic design frameworks (for example in \cite{romera2024mathematical,liu2024evolutionEoH,liu2025seoh,ye2025largeLLMLNS,zhang2025llmInstSpecHH,2024efficientHERCULES,huang2025calm}), we argue that the true value of automated heuristic design lies in domains where few heuristics exist or where established ones fail to perform effectively for specific instance configurations. The UPMP exemplifies the former case, as only a limited number of handcrafted heuristics are available. To represent the latter case, we study large sliding puzzles of size $20\times20$, for which pattern databases based state-of-the-art heuristics  \cite{culberson1998pattern,KORF20029,felner2004additive} become impractical due to the computational cost of pattern precomputation.
\par
Recent work showed that providing additional problem context can significantly enhance the quality of LLM-generated constructive heuristics for the UPMP, particularly for smaller LLMs \cite{bomer2025leveraging}.
We refer to this augmentation as \textbf{Problem - Contextual Evolution of Heuristics} (\texttt{P-CEoH}).
However, the main limitation of \texttt{P-CEoH} is the problem context still depends on the human formulating.
To address this limitation, we newly introduce \textbf{Algorithmic - Contextual Evolution of Heuristics (\texttt{A-CEoH})} that embeds key elements of the \texttt{A*} algorithm code into the prompt to guide the LLM toward generating heuristics that align with \texttt{A*}’s search dynamics.
The core structure of the \texttt{A*} is problem-agnostic and hence easily transferable.
\texttt{A-CEoH} aims to leverage heuristic generation by providing problem insights through code.
We are building on the \textbf{Evolution of Heuristics (\texttt{EoH})} by \cite{liu2024evolutionEoH} to investigate the new \texttt{A-CEoH}, the adapted \texttt{P-CEoH} \cite{bomer2025leveraging}, and a combination of both augmentations named \texttt{PA-CEoH} for the generation of \texttt{A*} guiding-heuristic for the UPMP and SPP. 
The main contributions of this paper are:
\begin{itemize}
    \item \textbf{New application field:} We demonstrate the potential of LLM-based automated heuristic design to evolve an \texttt{A*} guiding-heuristic 
    for a niche optimization problem and a well-studied optimization problem.
    \item \textbf{Algorithmic - Contextual Evolution of Heuristic:} We show that including algorithmic context in prompts significantly improves the performance of generated heuristics.
    \item \textbf{Performance boost for small models:} We illustrate that algorithmic context enables small, locally deployed LLMs to outperform large, general-purpose LLMs.
    \item \textbf{Outperforming handcrafted heuristics:} We compare the LLM-generated heuristics with hand-crafted heuristics from the literature and show the superiority of the LLM-generated heuristics for the target instance configuration.
\end{itemize}

\par
The rest of the paper is organized as follows. 
Section \ref{sec: combinatorial optimization problems} briefly introduces the two studied combinatorial optimization problems. 
Section \ref{sec: related work} provides an overview of solution approaches to the covered optimization problems, as well as automated heuristic design approaches using LLMs.
Section \ref{sec:Integrating Problem and Algorithmic Context into Evolutionary Heuristic Design} presents the evolutionary procedure and the proposed contextual augmentations. Section \ref{sec: Computational Experiments} presents the results of computational experiments. Finally, Section \ref{sec: Conclusion} summarizes the findings and outlines future research directions.

\section{Unit-load Pre-marshalling and Sliding Puzzle Problem}
\label{sec: combinatorial optimization problems}
The Unit-load Pre-marshalling Problem (UPMP), introduced in 2023 by \cite{pfrommer2023solving}, is about sorting unit loads in a block-stacking warehouse using autonomous robots. The goal is to sort unit loads by priority so that high-priority items can be accessed without being blocked with the minimal number of moves. A unit load is considered blocking if it prevents access to a higher-priority unit load, for example one that needs to be retrieved sooner.
During pre-marshalling, no new unit loads enter or leave.
\par
\begin{figure}
    \centering
    \includegraphics[width=1\linewidth]{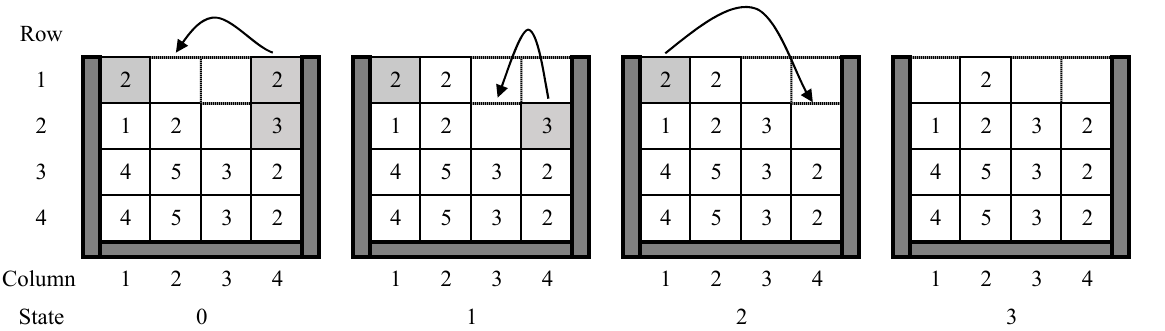}
    \caption{Example sequence of moves to solve an unit-load pre-marshalling problem instance by \cite{bomer2025leveraging}. Top-down view of a single-bay. Unit loads can be accessed from the north direction only. }
    \label{fig:instance_north_access_example}
\end{figure}

Figure \ref{fig:instance_north_access_example} shows a one-tier single-bay warehouse from a top-down view, where unit loads are accessed only from the north. In the initial state (0), three unit loads block access to higher-priority ones. After three moves (state 3), all blockages are resolved.

The Sliding Puzzle Problem (SPP) is a classic combinatorial optimization problem that has served as a benchmark for search algorithms since the 1980s \cite{KORF198597}. It involves rearranging tiles on a grid into a specified goal configuration by sliding them into an adjacent empty space. The objective is to reach the target arrangement with the minimal number of moves, where each move shifts one tile horizontally or vertically into the empty cell, changing the puzzle’s overall configuration.
\par
\begin{figure}
\centering
\includegraphics[width=1\linewidth]{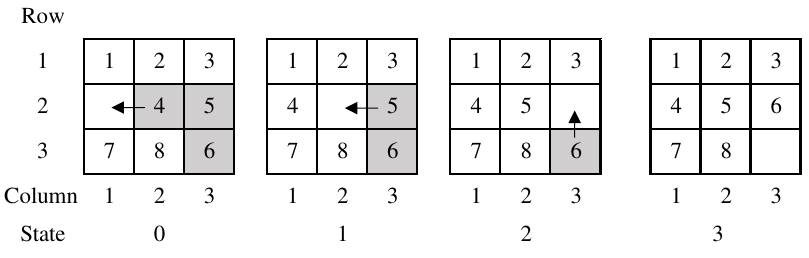}
\caption{Example sequence of moves to solve a $3\times3$ sliding puzzle. Tiles are moved into the empty cell to reach the goal configuration.}
\label{fig:sliding_puzzle_example}
\end{figure}

Figure \ref{fig:sliding_puzzle_example} shows a $3\times3$ sliding puzzle (also called 8-puzzle). In the initial state (0), three tiles are misplaced. By sliding tiles into the empty position, the correct order is gradually restored. After three moves (state 3), the puzzle reaches its solved configuration.

\section{Related Work}
\label{sec: related work}
In this section, we first present related work on pre-marshalling problems and the sliding puzzle problem. Following this, we present studies employing LLMs in an evolutionary framework to develop heuristics for combinatorial optimization problems.

\subsection{Related Work on Pre-marshalling}
For the UPMP in single-bay block-stacking warehouses, \cite{pfrommer2023solving} propose a two-stage solution approach. In the first stage, a network flow formulation is employed to decompose the warehouse bay into virtual lanes. In the second stage, an \texttt{A*} algorithm determines a solution that minimizes the total number of moves.
Building upon this work, \cite{BOMER2026107359sorting} extend the approach to multi-bay warehouse environments, aiming to minimize the loaded travel distance while maintaining the minimal number of reshuffling moves. Subsequently, \cite{bomer2024sorting,BOMER2025508CPsort} further enhance this approach by introducing a sequential scheme that allocates the moves generated by the tree search to multiple robots. This allocation considers move dependencies and is optimized through a mixed-integer programming formulation.
\par
The pre-marshalling of unit loads in block-stacking warehouses is conceptually related to the Container Pre-Marshalling Problem (CPMP) in maritime logistics, where containers within a bay must be rearranged until all target stacks are free of blockages. Among the solution strategies proposed for the CPMP—including greedy heuristics \cite{exposito2012pre}, dynamic programming \cite{prandtstetter2013dynamic}, integer programming \cite{lee2007optimization,parreno2019integer}, and constraint programming \cite{rendl2013constraint,jimenez2023constraint} tree search approaches have proven particularly effective.
Tree search methods \cite{bortfeldt2012tree,wang2024policy,hottung2020deep} systematically explore the solution space by branching on possible moves and applying heuristic or learned guidance to prune suboptimal states. 
\par
Heuristic and exact methods have been developed for both the UPMP and CPMP to minimize the number of moves; however, their design typically demands substantial algorithmic and domain-specific expertise.

\subsection{Related Work on Sliding Puzzles}
The SSP, including the 8- and 15-puzzle, has long served as a key benchmark for heuristic search, and finding optimal solutions for the general $n \times n$ case is NP-hard \cite{ratner1990n2}. Research on the 15-puzzle accelerated with the work of Korf \cite{KORF198597}, who showed that iterative-deepening-A*, combined with the Manhattan distance heuristic, was capable of producing the optimal solutions to random puzzle instances.
The authors of \cite{HANSSON1992207} introduced the linear conflict heuristic through a systematic method for generating admissible enhancements to Manhattan distance. A recent work \cite{hasan2023fifteen} proposed a human-designed hybrid heuristic—combining Manhattan distance, linear conflicts, and walking distance—within a bidirectional \texttt{A*} framework, yielding substantial reductions in generated states.
Pattern databases (PDBs) further advanced performance. The authors of \cite{culberson1998pattern} demonstrated large speed-ups using disjoint PDBs, enabling optimal solutions to the 24-puzzle. Additive heuristics \cite{KORF20029} allowed multiple PDBs to be combined while preserving admissibility, and later work \cite{felner2004additive} introduced static and dynamic partitioning strategies that produced state-of-the-art admissible heuristics.
\par
Although PDB-based methods dominate standard puzzle sizes, their memory and preprocessing demands make them infeasible for very large configurations such as the $20 \times 20$ puzzles considered in this work.

\subsection{Automated Heuristic Design with LLMs}
The development of LLM-based automated heuristic design was initiated by Google DeepMind's FunSearch \cite{romera2024mathematical} framework, and a range of subsequent approaches have built on its core idea. This section provides an overview:
\textbf{FunSearch} \cite{romera2024mathematical} integrates a pre-trained LLM with evolutionary search for algorithm discovery. Starting from simple hand-written heuristic code, the LLM generates new program code whose performance on benchmark instances is evaluated and stored.
\textbf{EoH} \cite{liu2024evolutionEoH} represents heuristics as both natural-language thoughts and executable code, refining both in parallel through a prompting scheme with two exploration and three modification strategies. It outperforms manual heuristics and FunSearch while requiring far fewer LLM queries (2,000 vs. one million for oBPP).
\textbf{P-CEoH} \cite{bomer2025leveraging} extends EoH with detailed, problem-specific contextual information to support constructive heuristic generation for niche combinatorial problems. Applied to the UPMP, it shows that smaller models like \texttt{Qwen2.5-Coder-32B} can generate heuristics outperforming larger models and those from standard EoH.
\textbf{ReEvo} \cite{ye2024reevo} augments evolutionary search with reflective reasoning. Its five-stage workflow—selection, short-term reflection, crossover, long-term reflection, and elitist mutation—uses repeated reflection to guide heuristic generation and consolidate insights from parent performance.
\textbf{HSEvo} \cite{dat2024hsevo} combines evolutionary operators with harmony search and a reflection mechanism (flash reflection). The authors aim to support diversity and tune elitist heuristic parameters via harmony search.
\textbf{Hercules} \cite{2024efficientHERCULES} enhances prompt efficiency via a reflection mechanism (Core Abstraction Prompting), which extracts key structure from elite heuristics, and reduces evaluation time via Performance Prediction Prompting, which estimates the heuristic fitness using semantic similarity with curated examples. 
\textbf{LLM-LNS} \cite{ye2025largeLLMLNS} applies LLM-driven heuristic evolution to Mixed Integer Linear Programming through Large Neighborhood Search. An inner layer evolves thoughts and code, while an outer layer adjusts prompts to maintain diversity.
\par
\par
\textbf{LLaMEA} \cite{liu2024largeLMEA} uses an LLM-guided evolutionary process to generate full metaheuristic algorithms for continuous black-box optimization. The LLM iteratively refines or redesigns the best algorithm using performance feedback, producing methods that outperform state-of-the-art metaheuristics. 
The domain-specific extension in \cite{yin2025optimizingllamea} applies LLaMEA to photonics using structured, domain-aware prompts and varied evolutionary strategies. Combining LLM-generated problem descriptions with algorithmic insights improves performance on two of three tested continuous optimization benchmarks.
\par
These studies demonstrate that LLMs can generate effective heuristics for combinatorial optimization. Since FunSearch and \texttt{EoH}, numerous extensions have added mechanisms such as reflection and diversity control. Because our work isolates the effects of prompt augmentation, we use \texttt{EoH} as our base framework: it offers strong prompt efficiency, a modular design, and ensures comparability with prior work such as \texttt{P-CEoH}\cite{bomer2025leveraging}.

\section{Integrating Algorithmic Context into Heuristic Design}
\label{sec:Integrating Problem and Algorithmic Context into Evolutionary Heuristic Design}
This section describes the adapted evolutionary framework Evolution of Heuristics (\texttt{EoH}) by \cite{liu2024evolutionEoH} and the adapted Problem - Contextual Evolution of Heuristics (\texttt{P-CEoH}) by \cite{bomer2025leveraging} briefly. Further, we explain the newly introduced prompt augmentation strategy Algorithmic - Contextual Evolution of Heuristics (\texttt{A-CEoH}). 
For illustrative examples of the prompt template and prompt augmentation elements, we refer to our \href{https://github.com/tb-git-tud/a-ceoh-evolution-of-heuristics}{repository}
\footnote{
\href{https://github.com/tb-git-tud/a-ceoh-evolution-of-heuristics?tab=readme-ov-file}{https://github.com/tb-git-tud/a-ceoh-evolution-of-heuristics} 
}.
Table \ref{tab:ceoh_components} shows the core components of each experimental setup. All versions adapt the evolutionary procedure to evolve heuristics in the form of Python code and an underlying thought. \texttt{P-CEoH} adds an additional problem description as described in \cite{bomer2025leveraging}. \texttt{A-EoH} instead adds an algorithmic context to leverage in-context learning and the development of a more targeted evolution for the algorithmic environment. \texttt{PA-CEoH} includes both: the additional problem description and algorithmic context to the prompt.

\begin{table}[!h]
\centering
\caption{Core components included in \texttt{EoH}, \texttt{P-CEoH}, \texttt{P-CEoH}, and \texttt{PA-CEoH}.}
\label{tab:ceoh_components}
\renewcommand{\arraystretch}{1.2}
\begin{tabular}{@{\extracolsep{3pt}}lcccc@{}}
\toprule
Component            & \texttt{EoH} & \texttt{P-CEoH} & \texttt{A-CEoH} &  \texttt{PA-CEoH} \\
\midrule
Evolutionary procedure & \checkmark   & \checkmark    & \checkmark &  \checkmark       \\
Heuristic (code + thought) & \checkmark   & \checkmark    & \checkmark & \checkmark      \\
Additional problem description        &             & \checkmark    &   & \checkmark      \\
Algorithmic description       &             &          & \checkmark   & \checkmark       \\
\bottomrule
\end{tabular}
\end{table}

\begin{figure}
    \centering
    \includegraphics[width=1\linewidth]{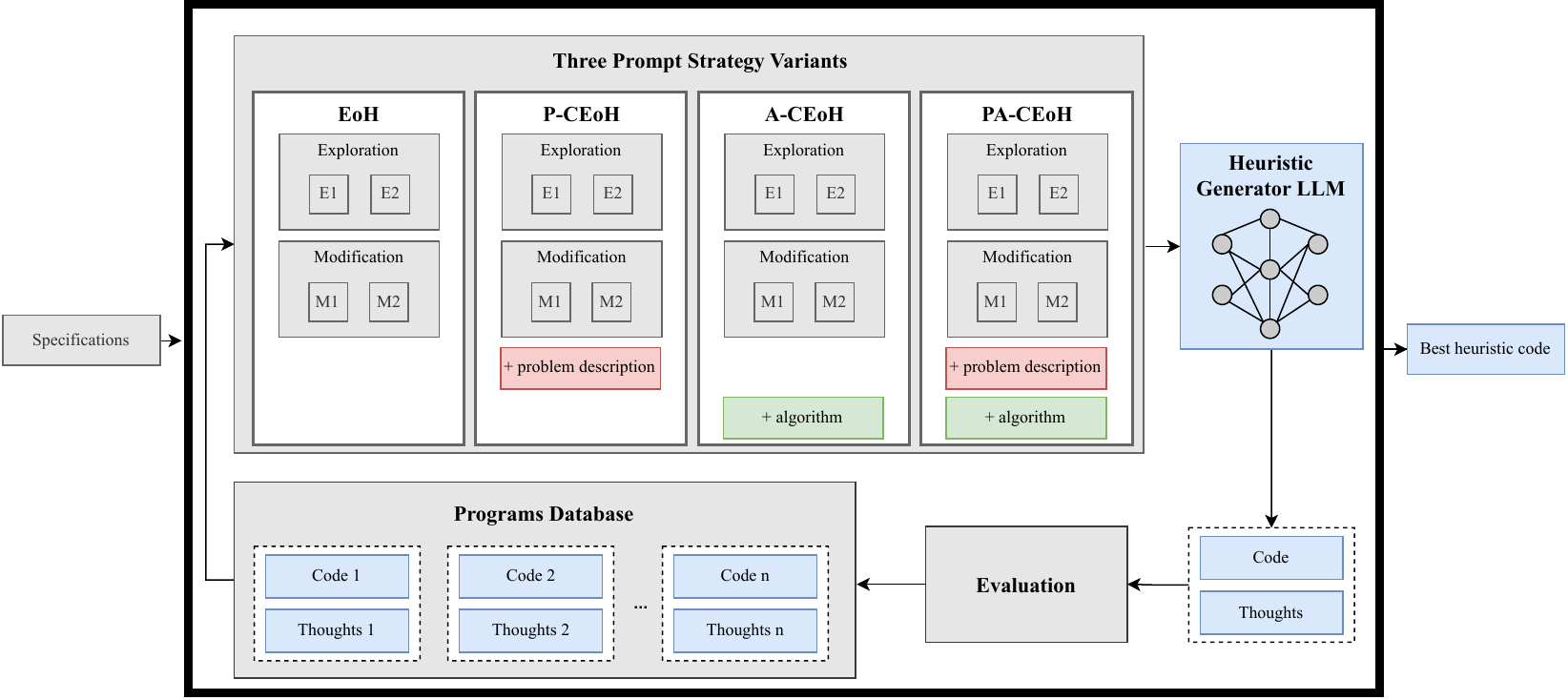}
    \caption{The \texttt{EoH} framework (grey and blue) evolves code and thoughts. The prompt augmentation components additional problem description (red) and algorithmic context (green) aim to support the evolutionary procedure.}
    \label{fig:augmentation_concepts}
\end{figure}

The Figure \ref{fig:augmentation_concepts} shows the evolutionary \texttt{EoH} framework and the prompt strategy variations we discuss in this paper. The \textcolor{red}{additional problem description} and the  \textcolor{mydarkgreen}{algorithmic context} are highlighted.
The original framework is grey and blue in Figure \ref{fig:augmentation_concepts}.

\paragraph{Evolutionary Procedure.}
The evolutionary framework follows an iterative process that constructs and refines a collection of heuristics over multiple generations. 
Each \textit{heuristic} $h \in \mathcal{H} = \{1, 2, \dots, |\mathcal{H}|\}$ consists of two components: executable \textit{program code} and accompanying \textit{thought}. 
The program code is implemented as a Python function with a clearly defined input--output interface, while the thought captures the conceptual reasoning expressed by the LLM during heuristic creation.
\par
The procedure progresses through a sequence of generations $\mathcal{G} = \{1, 2, \dots, |\mathcal{G}|\}$. 
The population of heuristics in a given generation $g \in \mathcal{G}$ is represented as $\mathcal{P}_{g} = \{1, 2, \dots, |\mathcal{P}_{g}|\}$. 
Within each generation, a set of prompt strategies $\mathcal{S} = \{1, 2, \dots, |\mathcal{S}|\}$ is employed to produce the next population from the heuristics of the preceding generation using a pre-trained LLM. 
Every strategy $s \in \mathcal{S}$ is applied $\bar{r}$ times, yielding a total of $|\mathcal{S}| \cdot \bar{r}$ new heuristics for generation $g$.
\par
Each newly produced heuristic $h$ is evaluated on a collection of problem instances $\mathcal{I} = \{1, 2, \dots, |\mathcal{I}|\}$, where its performance is quantified by a fitness score $f^{\mathcal{I}}(h)$. 
After evaluation, $h$ is added to the current population $\mathcal{P}_{g}$, and the process continues until all heuristics for the generation have been created. 
At the end of each generation, the top-performing $\bar{n}$ heuristics are retained and carried over to the next generation. 
An initialization phase generates the initial population $\mathcal{P}_{0}$ by requesting an initial prompt $2 \cdot \bar{n}$ times.

\paragraph{Heuristic Prompt Strategies}
This work employs the prompt strategies by \cite{liu2024evolutionEoH}, including the initialization prompt I1, two exploration prompts (E1 and E2), and two modification prompts (M1 and M2). 
The following summarizes the functionality of each strategy.

\begin{itemize}[label={}]
  \item \textbf{I1:} Formulate a heuristic aimed at solving the given optimization task.
  \item \textbf{E1:} Develop a heuristic that differs substantially from $\bar{p}$ parent heuristics sampled from the current population.
  \item \textbf{E2:} Generate a heuristic that retains the core conceptual idea of $\bar{p}$ parent heuristics while offering a new variation.
  \item \textbf{M1:} Revise an existing parent heuristic from the population to enhance its effectiveness.
  \item \textbf{M2:} Modify the parameter configuration of a parent heuristic from the population to improve its overall performance.
\end{itemize}

\par
Each prompt consists of these elements: 
(1) task description, 
(2) algorithmic context,
(3) additional problem description, 
(3) parent heuristic(s) (not in I1), 
(4) strategy-specific output instructions, 
and (5) additional instructions. 
We detail each component in the following.
\par
The \textbf{(1) task description} informs the LLM about the optimization problem and how the heuristic will be used in a bigger context. 
The \textcolor{mydarkgreen}{\textbf{(2) algorithmic context}} provides the LLM with the precise algorithmic environment in which its heuristic is applied. Figure \ref{fig:astar_algorithmic_context} shows the algorithmic context provided for the SPP as an example.
It begins by presenting the complete \texttt{A*} driver procedure, which defines the overall search logic, including the initialization of the open list, node expansion, and goal detection. 
This code shows how the heuristic value \( h(n) \), computed via the LLM-generated function \texttt{score\_state(state)}, is combined with the path cost \( g(n) \) to form the evaluation function \( f(n) = g(n) + h(n) \). 
After the driver, the context includes the core methods that define the problem’s search space and objective structure: 
(a) \texttt{is\_goal()} specifies the objective of the search; 
(b) \texttt{get\_neighbors()} generates successor states and thus determines how the search tree expands; 
(c) \texttt{reconstruct\_path()} rebuilds the sequence of actions leading to a found goal; and 
(d) \texttt{get\_objective\_value()} quantifies solution quality according to the task’s optimization criterion. 
Together, these components convey how the heuristic integrates into the \texttt{A*}  search process, how it influences node prioritization, and how the resulting solution is evaluated. 
Because this structure is domain-agnostic, the same prompt format can be used for different optimization problems simply by substituting the corresponding functions.
The \textcolor{red}{\textbf{(3) additional problem description}} (adapted from \cite{bomer2025leveraging}) enhances the prompt with contextual details about the optimization setting, supporting stronger in-context understanding by the LLM. 
It first outlines the expected \textit{structure of the input data} that the heuristic will process. 
Then, it clarifies the \textit{interpretation} of the individual components within the problem domain.  
Next, it specifies problem-specific \textit{constraints}. 
Finally, it supplements the description with \textit{illustrative examples} of both input and output data. 
The \textbf{(4) parent heuristic(s)} are supplied as code snippets accompanied by their underlying thoughts. 
These examples act as few-shot demonstrations that guide the LLM’s generation process, helping it identify structural and conceptual patterns from prior heuristics. 
The \textbf{(5) strategy-specific output instructions} determine how the LLM should formulate its response. 
They define the expected structure of the output—both the heuristic’s reasoning (\textit{thought}) and its executable program code with well-defined inputs and outputs—tailored to the respective prompt strategy. 
Finally, the \textbf{(6) additional instructions} impose essential implementation and stylistic constraints. 
To ensure logical coherence, self-consistency is required so that the generated Python function aligns with its corresponding thought \cite{min2023beyond}. 

\lstset{
    language=Python,
    basicstyle=\ttfamily\tiny,
    keywordstyle=\bfseries\color{blue},
    commentstyle=\itshape\color{gray},
    stringstyle=\color{black},
    numbers=left,
    numberstyle=\tiny,
    stepnumber=1,
    frame=single,
    tabsize=2,
    showstringspaces=false,
    xleftmargin=0pt,
    xrightmargin=0pt,
    aboveskip=1em,
    belowskip=1em
}

\begin{figure}[]
\centering
\begin{minipage}{1\textwidth}
\tiny
\begin{lstlisting}[language=Python, label={lst:astar_core}]
def astar_puzzle_core(heuristics, start_puzzle):
    open_list = []
    visited = set()
    evaluated_nodes = 0
    counter = itertools.count()

    root = PuzzleNode(start_puzzle, g=0, heuristic_fn=heuristics.score_state)
    heapq.heappush(open_list, (root.f, next(counter), root))
    visited.add(root.serialize())
    start = time.monotonic()

    while open_list:
        # Timeout / node cap check
        if ((time.monotonic() - start) > TIMEOUT_SECONDS or 
            evaluated_nodes > MAX_EVALUATED_NODES):
            return return_result(False)

        _, _, current = heapq.heappop(open_list)

        if current.is_goal():
            return return_result(True)

        for neighbor in current.get_neighbors(heuristics.score_state):
            state = neighbor.serialize()
            if state in visited:
                continue
            evaluated_nodes += 1
            visited.add(state)
            heapq.heappush(open_list, (neighbor.f, next(counter), neighbor))

def is_goal(self):
    flat = [num for row in self.puzzle for num in row]
    return flat == list(range(1, self.N * self.N)) + [0]

def get_neighbors(self, heuristic_fn):
    neighbors = []
    r, c = self.find_blank()
    directions = {(-1, 0): 'U', (1, 0): 'D', (0, -1): 'L', (0, 1): 'R'}
    for (dr, dc), move in directions.items():
        nr, nc = r + dr, c + dc
        if 0 <= nr < self.N and 0 <= nc < self.N:
            new_puzzle = deepcopy(self.puzzle)
            new_puzzle[r][c], new_puzzle[nr][nc] = new_puzzle[nr][nc], new_puzzle[r][c]
            neighbors.append(PuzzleNode(new_puzzle, self.g + 1, heuristic_fn, parent=self, move=move))
    return neighbors

def reconstruct_path(node):
    path = []
    while node.parent is not None:
        path.append((node.move, node.puzzle))
        node = node.parent
    return list(reversed(path))
    
def get_objective_value(self):
    if not self.is_goal():
        return MAX_MOVES
    return len(self.reconstruct_path(self))
\end{lstlisting}
\end{minipage}
\caption{Algorithmic context for the SPP.}
\label{fig:astar_algorithmic_context}
\end{figure}

\section{Computational Experiments}
\label{sec: Computational Experiments}
We conducted extensive experiments to show the model-specific effects of \texttt{A-CEoH}, \texttt{P-CEoH}, and \texttt{PA-CEoH}.

\paragraph{Evolutionary Parameters.}
Each framework version (\texttt{EoH}, \texttt{P-CEoH}, \texttt{A-CEoH}, and P\texttt{A-CEoH}) is executed ten times per LLM for the UPMP and five times per LLM for the SPP. 
All versions use the initialization prompt I1, called $40$ times, with the $20$ best heuristics forming the initial population. 
Heuristics evolve over $20$ generations using prompt strategies E1, E2, M1, and M2, each invoked $\bar{r}=20$ times per generation, resulting in $20 \cdot 4 \cdot 20 = 1600$ prompts per run. 
The number of parent heuristics for E1 and E2 is set to $p=5$. 

\paragraph{Evaluation Parameters.}

For the UPMP, we use a 60\,s time limit and a node limit of 100{,}000 per instance; unsolved instances are penalized with $m^{max}=100$. Evaluation instances use a single-bay $5\times5$ layout (one tier, north access), five priority classes, and a 60\,\% fill rate.
For the SPP, we apply a 60\,s time limit and a node limit of 1{,}000{,}000; unsolved instances receive $m^{max}=200$. We consider large $20\times20$ puzzles generated by 200 random moves, a scale at which pattern-database heuristics are impractical (published with code).
For both problems, we use ten training instances (seeds 0--9). All experiments were conducted on an AMD EPYC 7401P with 64\,GB RAM.

\paragraph{Fitness Calculation}

Each heuristic $h$ is evaluated over a set of instances $\mathcal{I}$. Let $m_i$ denote the number of moves required to solve instance $i$, and $m_i^{lb}$ a lower bound on the required moves. For the UPMP, lower bounds follow \cite{pfrommer2023solving}; for the SPP, we use the number of misplaced tiles, which suffices for qualitative comparisons. If $h$ fails to solve instance $i$, we set $m_i = m^{max} \gg m_i^{lb}$. The fitness is defined as the average relative deviation from the lower bound (Eq.~\ref{eq:fitness_funtion}) and is minimized. Since $m_i^{lb}$ is not necessarily optimal, a fitness of zero is generally unattainable.

\begin{equation}
\label{eq:fitness_funtion}
f^{\mathcal{I}}(h) = 
\frac
{
1
}
{
|\mathcal{I}|
}
\displaystyle\sum_{i \in \mathcal{I}}\frac{m_{i}-m^{lb}_{i}}{m^{lb}_{i}}
\end{equation}

\begin{figure}[htbp]
    \centering
    \begin{subfigure}{0.48\linewidth}
        \centering
        \includegraphics[width=\linewidth]{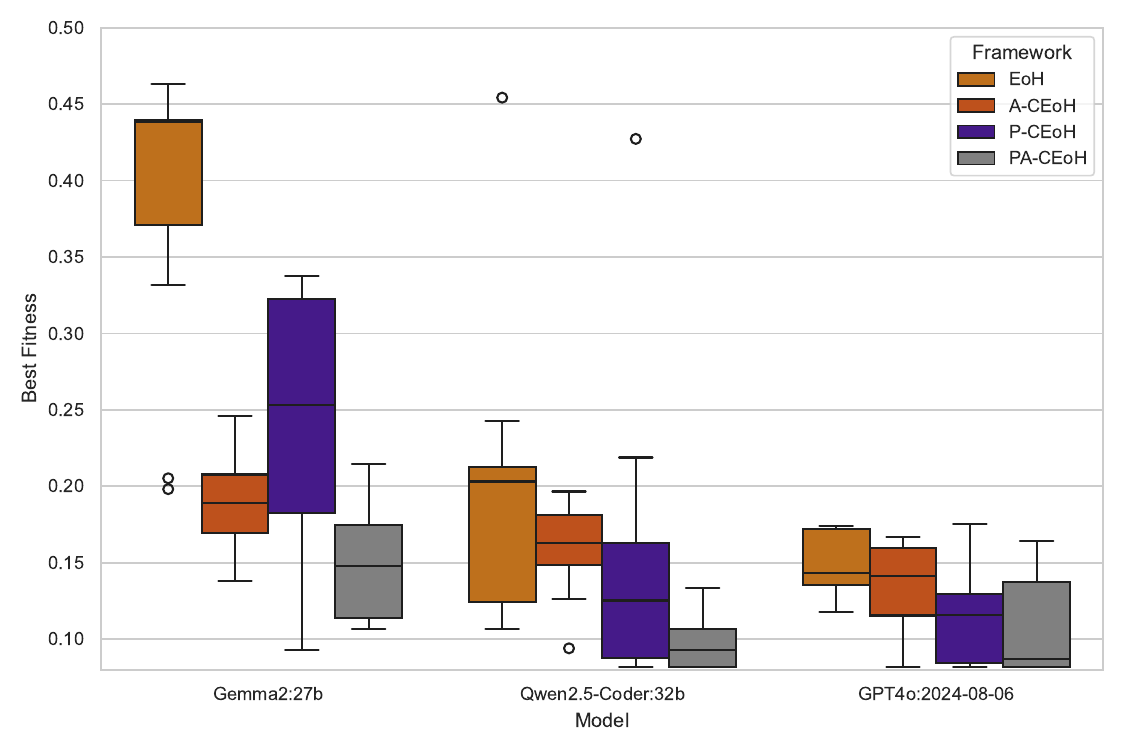}
        \caption{UPMP}
        \label{fig:best_heuristics_upmp}
    \end{subfigure}
    \hfill
    \begin{subfigure}{0.48\linewidth}
        \centering
        \includegraphics[width=\linewidth]{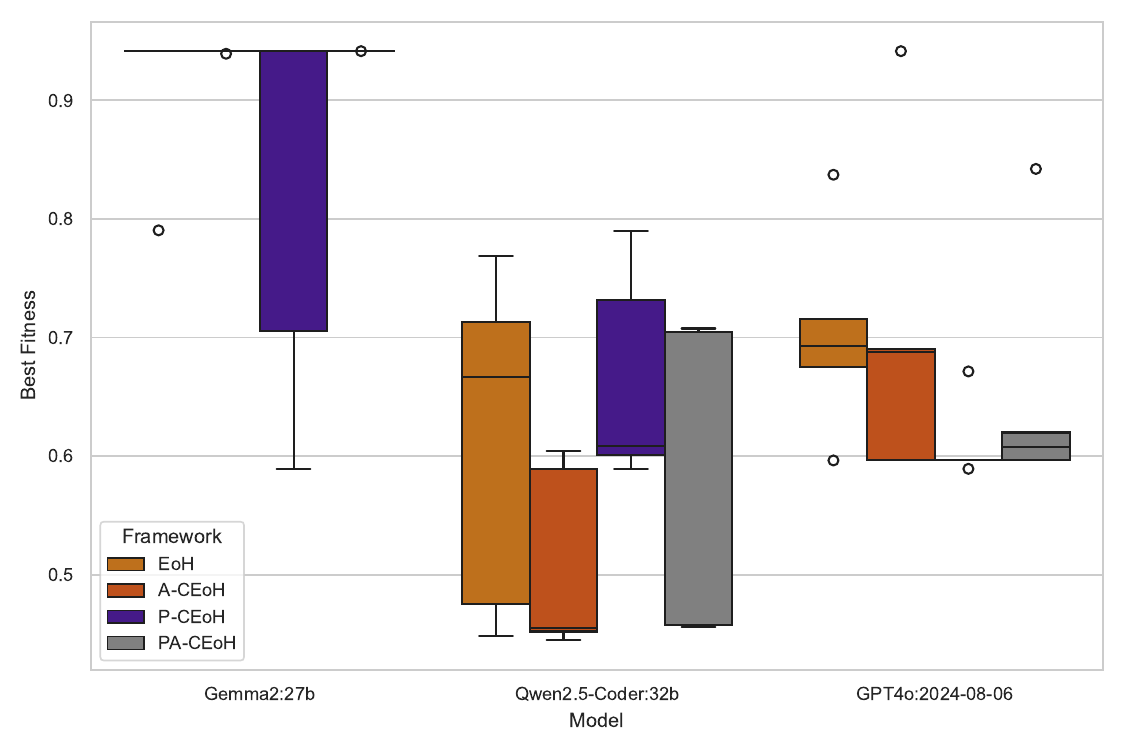}
        \caption{SPP}
        \label{fig:best_heuristics_spp}
    \end{subfigure}
    \caption{Best heuristic fitness in each experimental run for each model for the UPMP and SPP. Lower values indicate better performance. The fitness scale for the UPMP plot is truncated at 0.5 and at 1 for the SPP. Each boxplot reports ten runs for the UPMP and five runs for the SPP.}
    \label{fig:best_heuristics_combined}
\end{figure}

\paragraph{Effect of Context.}
Figure \ref{fig:best_heuristics_combined} illustrates the distribution of fitness values for the best heuristic identified by each LLM.
The detailed results for the UPMP are shown in Figure \ref{fig:best_heuristics_upmp}, and those for the SPP in Figure \ref{fig:best_heuristics_spp}.
For the UPMP, the newly introduced \texttt{A-CEoH} consistently outperforms the \texttt{EoH} baseline in both the best-found heuristic and the median fitness across all evaluated models.
The adapted \texttt{P-CEoH} achieves strong performance for the \texttt{Qwen2.5-Coder:32b} and \texttt{GPT4o:2024-08-06}, although it exhibits instability for the \texttt{Gemma2:27b}.
The combined variant, \texttt{PA-CEoH}, delivers the best overall results, demonstrating that integrating both prompt augmentations provides complementary benefits across all tested models. Multiple runs acquire the best reachable fitness value of $0.0815$.
\par
For the SPP, the \texttt{Qwen2.5-Coder:32b} benefits most from the newly introduced algorithmic context in \texttt{A-CEoH}. 
The best SPP heuristic was generated by \texttt{Qwen2.5-Coder:32b} in the \texttt{A-CEoH} setup with a fitness value of $0.445$.
In contrast, the \texttt{P-CEoH} augmentation setup does not improve heuristic quality for the \texttt{Qwen2.5-Coder:32b}.
The larger \texttt{GPT4o:2024-08-06} achieves moderate improvements in performance through prompt augmentation but fails to surpass a fitness value of 0.6, suggesting convergence to a local optimum.
\texttt{Gemma2:27b} struggles to produce competitive heuristics for the SPP, except in a few isolated runs in the \texttt{P-CEoH} setup.
Overall, the smaller, coding-oriented \texttt{Qwen2.5-Coder:32b} generates significantly stronger heuristics than the larger \texttt{GPT4o:2024-08-06}. Especially with the new \texttt{A-CEoH} prompt context.

\begin{figure}
    \centering
    \includegraphics[width=1\linewidth]{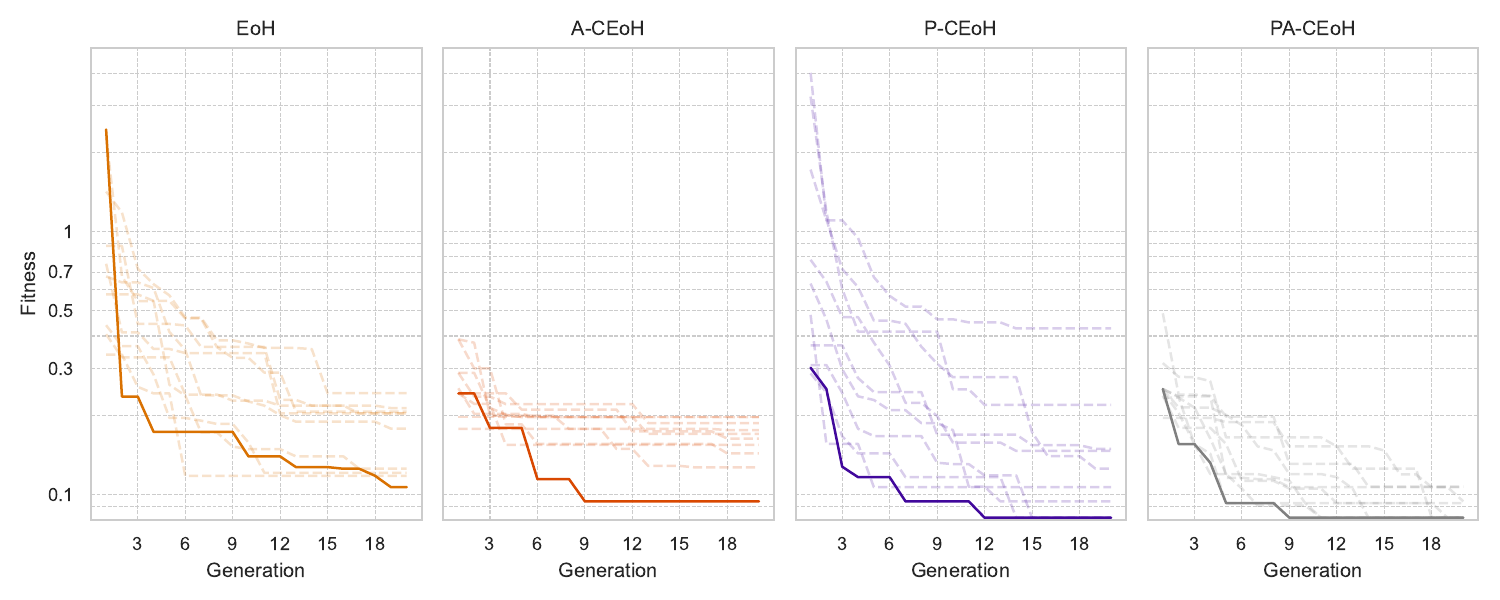}
    \caption{Fitness of the best found heuristic by \texttt{Qwen2.5-Coder:32b} for the UPMP across generations for each experiment run. The best run for each experimental setup is shown with opacity. Lower values indicate better performance.}
    \label{fig:qwen_fiteness_over_generations}
\end{figure}

Figure \ref{fig:qwen_fiteness_over_generations} illustrates the convergence of fitness values across generations for each prompt augmentation strategy applied to the \texttt{Qwen2.5-Coder:32b} model for the UPMP task. The \texttt{A-CEoH} approach enables the LLM to generate reasonably good heuristics early in the evolutionary process but exhibits difficulty converging toward better fitness values.
In contrast, the experimental runs under the \texttt{P-CEoH} setting begin similarly to the base \texttt{EoH} configuration, but the inclusion of additional problem context allows the model to identify improvements more reliably. 
Finally, the combined \texttt{PA-CEoH} strategy leverages the strengths of both augmentations—adopting the \texttt{A-CEoH}’s ability to produce good heuristics early on while benefiting from the \texttt{P-CEoH}’s capacity to achieve stronger overall improvements.

\paragraph{Comparison with Human-designed Heuristics.}
We compare the best LLM-generated heuristics with \texttt{A*} guiding-heuristics proposed in prior work.
To ensure a fair evaluation, the heuristics from the literature are adapted and integrated into our \texttt{A*} implementation so that only the guiding-heuristic component differs. We limit the solution time to 600 seconds. Fitness and solution time are reported on average for the solved instances.
\par
Table~\ref{tab:heuristic_benchmark_astar_UPMP} presents the comparison between the two best heuristics generated for the UPMP and the optimal heuristic introduced by \cite{pfrommer2023solving}. For the training instances (seeds 0–9), the LLM-generated heuristics achieve optimal fitness values while solving the instances faster. On an additional 30 test instances (seeds 10–39), the LLM-generated heuristics successfully solve cases that the optimal implementation fails to solve, while remaining near-optimal in performance.

\begin{table}[]
\centering
\caption{Comparison of the best LLM-generated UPMP \texttt{A*}  guiding-heuristics with an optimal \texttt{A*}  implementation from the literature.}
\label{tab:heuristic_benchmark_astar_UPMP}
\begin{tabularx}{\textwidth}{XXXXXXXXXX} 
\toprule
&
\multicolumn{6}{c}{\texttt{P\texttt{A-CEoH}}} &
 \\
 \cmidrule{2-7}
&
\multicolumn{3}{c}{\texttt{Qwen2.5-Coder:32b}} &
\multicolumn{3}{c}{\texttt{GPT4o:2024-08-06}} &
\multicolumn{3}{c}{Optimal \texttt{A*}  \cite{pfrommer2023solving}} \\
\cmidrule{2-4} \cmidrule{5-7} \cmidrule{8-10}
seeds & solved & fitness & time & solved & fitness & time & solved & fitness & time \\
\midrule
0–9 & 10 & 0.0815 & 0.0888 & 10 & 0.0815 & 0.1953 & 10 & 0.0815 & 0.2773 \\
10–39 & 30 & 0.1388 & 0.151 & 30 & 0.1055 & 0.098 & 27 & 0.0968 & 0.44 \\
\bottomrule
\end{tabularx}
\end{table}
\par
Table \ref{tab:heuristic_benchmark_astar_SPP} compares the best LLM-generated heuristic for the SPP with a human-designed hybrid heuristic~\cite{hasan2023fifteen} and an optimal implementation based on Manhattan distance and linear conflicts \cite{HANSSON1992207}. The generated heuristic is able to solve instances that both benchmark heuristics cannot. 
Considering only the five instances solved by both the optimal approach and the LLM-generated heuristic for seeds 0–9, the LLM-generated heuristic achieves a fitness value of $0.281$, which is a $0.031$ gap to the optimal $0.252$ fitness value.
Moreover, the LLM-generated heuristic achieves significantly shorter runtime while maintaining fitness values comparable to the optimal approach for seeds 10–19.

\begin{table}[htpb]
\centering
\caption{Comparison of the best LLM-generated SPP \texttt{A*}  guiding-heuristics with a heuristic and an optimal \texttt{A*}  implementation from the literature.}
\label{tab:heuristic_benchmark_astar_SPP}
\begin{tabularx}{\textwidth}{XXXX XXX XXX} 
\toprule
& \multicolumn{3}{c}{\texttt{\texttt{A-CEoH}}} \\
\cmidrule{2-4}
&
\multicolumn{3}{c}{\texttt{Qwen2.5-Coder:32b}} &
\multicolumn{3}{c}{Hybrid Heuristic \texttt{A*}  \cite{hasan2023fifteen}} &
\multicolumn{3}{c}{Optimal \texttt{A*}  \cite{HANSSON1992207}} \\
\cmidrule{2-4} \cmidrule{5-7} \cmidrule{8-10}
seeds &solved & fitness & time & solved & fitness & time & solved & fitness & time \\
\midrule
0–9 & 10 & 0.445 & 17.417 & 5 & 0.539 & 56.587 & 5 & 0.252 & 137.55 \\
10–19 & 7 & 0.392 & 16.272 & 6 & 0.479 & 189.51 & 6 & 0.375 & 187.91 \\
\bottomrule
\end{tabularx}
\end{table}

\paragraph{Token Usage}
Table~\ref{tab:token_usage_UPMP_SPP} reports the mean input and output token counts per prompt for both problems using \texttt{Qwen2.5-Coder:32b}. Adding contextual information substantially increases input tokens (\texttt{PA-CEoH} $\approx \times 2 $) compared to \texttt{EoH} while generally reducing output tokens, indicating more focused generation.

\begin{table}[]
    \centering
    \caption{Mean number of input and output tokens per prompt for the \texttt{Qwen2.5-Coder:32b} for the UPMP and SPP.}
    \label{tab:token_usage_UPMP_SPP}
    \begin{tabular*}{\linewidth}{@{\extracolsep{\fill}}lrrrrrrrr} 
    \toprule
     & \multicolumn{2}{c}{\texttt{EoH}} & \multicolumn{2}{c}{\texttt{A-CEoH}} & \multicolumn{2}{c}{\texttt{P-CEoH}} & \multicolumn{2}{c}{\texttt{PA-CEoH}} \\
         \cmidrule{2-3}\cmidrule{4-5}\cmidrule{6-7}\cmidrule{8-9}
     & Input & Output & Input & Output & Input & Output & Input & Output \\
    \midrule
    UPMP & 1046 & 259 & 1642 & 178 & 1474 & 230 & 2223 & 215 \\ 
    SPP & 1477 & 401 & 2245 & 387 & 2823 & 378 & 3330 & 305 \\
    \bottomrule
    \end{tabular*}
\end{table}

\section{Conclusion}
\label{sec: Conclusion}
This study demonstrated that LLMs can autonomously design effective \texttt{A*} guiding-heuristics for complex combinatorial optimization problems. Building on the \texttt{EoH} framework, we introduced \texttt{A-CEoH}, a novel prompt augmentation strategy that embeds the \texttt{A*}  code structure directly into LLM prompts.
We compared \texttt{A-CEoH} with the \texttt{P-CEoH}, an adapted approach from the literature that enhances prompts with domain-specific problem context. Experiments on the UPMP and the SPP showed that incorporating algorithmic context in \texttt{A-CEoH} leads to a strong improvement in heuristic quality compared to the original \texttt{EoH} baseline.
When both augmentations were combined in PA-CEoH, the results improved further for the UPMP, demonstrating that algorithmic and problem-specific context provide complementary benefits for the studied niche optimization problem.
For the well-studied SPP, the performance of the \texttt{PA-CEoH} did not exceed that of the A-CEoH.
Across both problem domains, the smaller coding-oriented model \texttt{Qwen2.5-Coder:32b} performed on par with or better than the much larger \texttt{GPT4o:2024-08-06}, highlighting that well-structured contextual prompts can outweigh model size.
A comparison of the LLM-generated heuristics with human-designed \texttt{A*} guiding-heuristics from the literature showed superior performance of the LLM-generated heuristics.
These findings emphasize the importance of algorithm-aware prompt design and represent a step toward automated, context-sensitive heuristic discovery for constrained optimization tasks. 
Future work will extend the problem-agnostic idea of algorithmic context to additional combinatorial problems and to other algorithmic frameworks such as large neighborhood search.

%% file: sources.bib
@ARTICLE{hartAstar,
  author={Hart, Peter E. and Nilsson, Nils J. and Raphael, Bertram},
  journal={IEEE Transactions on Systems Science and Cybernetics}, 
  title={A Formal Basis for the Heuristic Determination of Minimum Cost Paths}, 
  year={1968},
  volume={4},
  number={2},
  pages={100-107},
  doi={10.1109/TSSC.1968.300136}}

@article{KORF198597,
title = {Depth-first iterative-deepening: An optimal admissible tree search},
journal = {Artificial Intelligence},
volume = {27},
number = {1},
pages = {97-109},
year = {1985},
issn = {0004-3702},
doi = {https://doi.org/10.1016/0004-3702(85)90084-0},
url = {https://www.sciencedirect.com/science/article/pii/0004370285900840},
author = {Richard E. Korf}
}

@article{hasan2023fifteen,
  title={The fifteen puzzle—A new approach through hybridizing three heuristics methods},
  author={Hasan, Dler O and Aladdin, Aso M and Talabani, Hardi Sabah and Rashid, Tarik Ahmed and Mirjalili, Seyedali},
  journal={Computers},
  volume={12},
  number={1},
  pages={11},
  year={2023},
  publisher={MDPI}
}

@article{KORF20029,
title = {Disjoint pattern database heuristics},
journal = {Artificial Intelligence},
volume = {134},
number = {1},
pages = {9-22},
year = {2002},
issn = {0004-3702},
doi = {https://doi.org/10.1016/S0004-3702(01)00092-3},
url = {https://www.sciencedirect.com/science/article/pii/S0004370201000923},
author = {Richard E. Korf and Ariel Felner}
}

@article{HANSSON1992207,
title = {Criticizing solutions to relaxed models yields powerful admissible heuristics},
journal = {Information Sciences},
volume = {63},
number = {3},
pages = {207-227},
year = {1992},
issn = {0020-0255},
doi = {https://doi.org/10.1016/0020-0255(92)90070-O},
url = {https://www.sciencedirect.com/science/article/pii/002002559290070O},
author = {Othar Hansson and Andrew Mayer and Moti Yung}
}

@article{liu2025seoh,
  title={Eoh-s: Evolution of heuristic set using llms for automated heuristic design},
  author={Liu, Fei and Liu, Yilu and Zhang, Qingfu and Tong, Xialiang and Yuan, Mingxuan},
  journal={arXiv preprint arXiv:2508.03082},
  year={2025}
}

@inproceedings{ye2025largeLLMLNS,
  title={Large Language Model-driven Large Neighborhood Search for Large-Scale MILP Problems},
  author={Ye, Huigen and Xu, Hua and Yan, An and Cheng, Yaoyang},
  booktitle={Forty-second International Conference on Machine Learning},
  year={2025}
}

@article{huang2025calm,
  title={Calm: Co-evolution of algorithms and language model for automatic heuristic design},
  author={Huang, Ziyao and Wu, Weiwei and Wu, Kui and Wang, Jianping and Lee, Wei-Bin},
  journal={arXiv preprint arXiv:2505.12285},
  year={2025}
}

@article{zhang2025llmInstSpecHH,
  title={Llm-driven instance-specific heuristic generation and selection},
  author={Zhang, Shaofeng and Liu, Shengcai and Lu, Ning and Wu, Jiahao and Liu, Ji and Ong, Yew-Soon and Tang, Ke},
  journal={arXiv preprint arXiv:2506.00490},
  year={2025}
}

@article{jimenez2023constraint,
  title={A constraint programming approach for the premarshalling problem},
  author={Jim{\'e}nez-Piqueras, Celia and Ruiz, Rub{\'e}n and Parre{\~n}o-Torres, Consuelo and Alvarez-Valdes, Ramon},
  journal={European Journal of Operational Research},
  volume={306},
  number={2},
  pages={668--678},
  year={2023},
  publisher={Elsevier},
  doi={10.1016/j.ejor.2022.07.042}
}

@article{pfrommer2023solving,
  title={Solving the unit-load pre-marshalling problem in block stacking storage systems with multiple access directions},
  author={Pfrommer, Jakob and Meyer, Anne and Tierney, Kevin},
  journal={European Journal of Operational Research},
  year={2023},
  publisher={Elsevier},
  doi ={10.1016/j.ejor.2023.08.044}
}

@article{lee2007optimization,
  title={An optimization model for the container pre-marshalling problem},
  author={Lee, Yusin and Hsu, Nai-Yun},
  journal={Computers \& operations research},
  volume={34},
  number={11},
  pages={3295--3313},
  year={2007},
  publisher={Elsevier},
  doi ={10.1016/j.cor.2005.12.006} 
}

@article{parreno2019integer,
  title={Integer programming models for the pre-marshalling problem},
  author={Parre{\~n}o-Torres, Consuelo and Alvarez-Valdes, Ramon and Ruiz, Rub{\'e}n},
  journal={European Journal of Operational Research},
  volume={274},
  number={1},
  pages={142--154},
  year={2019},
  publisher={Elsevier},
  doi= {10.1016/j.ejor.2018.09.048 }
}

@article{bortfeldt2012tree,
  title={A tree search procedure for the container pre-marshalling problem},
  author={Bortfeldt, Andreas and Forster, Florian},
  journal={European Journal of Operational Research},
  volume={217},
  number={3},
  pages={531--540},
  year={2012},
  publisher={Elsevier},
  doi={10.1016/j.ejor.2011.10.005} 
}

@article{prandtstetter2013dynamic,
  title={A dynamic programming based branch-and-bound algorithm for the container pre-marshalling problem},
  author={Prandtstetter, Matthias},
  journal={Technical repot, IT Austrian institute of technology},
  year={2013},
  doi={}
}

@inproceedings{rendl2013constraint,
  title={Constraint models for the container pre-marshaling problem},
  author={Rendl, Andrea and Prandtstetter, Matthias},
  booktitle={ModRef 2013: The Twelfth International Workshop on Constraint Modelling and Reformulation},
  year={2013},
  url={https://api.semanticscholar.org/CorpusID:43049743}
}

@article{exposito2012pre,
  title={Pre-marshalling problem: Heuristic solution method and instances generator},
  author={Exp{\'o}sito-Izquierdo, Christopher and Meli{\'a}n-Batista, Bel{\'e}n and Moreno-Vega, Marcos},
  journal={Expert Systems with Applications},
  volume={39},
  number={9},
  pages={8337--8349},
  year={2012},
  publisher={Elsevier}
}

@article{BOMER2025508CPsort,
title = {A Constraint Programming Approach for the Multi-Robot Multibay Unit Load Pre-marshalling Problem},
journal = {Procedia CIRP},
volume = {134},
pages = {508-513},
year = {2025},
note = {58th CIRP Conference on Manufacturing Systems 2025},
issn = {2212-8271},
doi = {https://doi.org/10.1016/j.procir.2025.02.151},
url = {https://www.sciencedirect.com/science/article/pii/S2212827125005359},
author = {Thomas Bömer and Max Disselnmeyer and Anne Meyer}
}

@article{BOMER2026107359sorting,
title = {Sorting multi–bay block stacking storage systems},
journal = {Computers \& Operations Research},
volume = {188},
pages = {107359},
year = {2026},
issn = {0305-0548},
doi = {https://doi.org/10.1016/j.cor.2025.107359},
url = {https://www.sciencedirect.com/science/article/pii/S0305054825003880},
author = {Thomas Bömer and Jakob Pfrommer and Daniyar Akizhanov and Anne Meyer}
}

@InProceedings{bomer2025leveraging,
author="B{\"o}mer, Thomas
and Koltermann, Nico
and Disselnmeyer, Max
and D{\"o}rr, Laura
and Meyer, Anne",
editor="Marcelloni, Francesco
and Madani, Kurosh
and van Stein, Niki
and Filipe, Joaquim",
title="Prompt-Augmentation for Evolving Heuristics for a Niche Optimization Problem",
booktitle="Computational Intelligence",
year="2026",
publisher="Springer Nature Switzerland",
address="Cham",
pages="214--235",
isbn="978-3-032-15632-7"
}

@inproceedings{liu2024evolutionEoH,
  title={Evolution of Heuristics: Towards Efficient Automatic Algorithm Design Using Large Language Model},
  author={Liu, Fei and Xialiang, Tong and Yuan, Mingxuan and Lin, Xi and Luo, Fu and Wang, Zhenkun and Lu, Zhichao and Zhang, Qingfu},
  booktitle={Forty-first International Conference on Machine Learning},
  year={2024}
}

@article{ye2024reevo,
  title={Reevo: Large language models as hyper-heuristics with reflective evolution},
  author={Ye, Haoran and Wang, Jiarui and Cao, Zhiguang and Song, Guojie},
  journal={arXiv preprint arXiv:2402.01145}  ,
  year={2024}
}

@inproceedings{2024efficientHERCULES,
author = {Wu, Xuan and Wang, Di and Wu, Chunguo and Wen, Lijie and Miao, Chunyan and Xiao, Yubin and Zhou, You},
title = {Efficient Heuristics Generation for Solving Combinatorial Optimization Problems Using Large Language Models},
year = {2025},
isbn = {9798400714542},
publisher = {Association for Computing Machinery},
address = {New York, NY, USA},
url = {https://doi.org/10.1145/3711896.3736923},
doi = {10.1145/3711896.3736923},
booktitle = {Proceedings of the 31st ACM SIGKDD Conference on Knowledge Discovery and Data Mining V.2},
pages = {3228–3239},
numpages = {12},
location = {Toronto ON, Canada},
series = {KDD '25}
}

@article{romera2024mathematical,
  title={Mathematical discoveries from program search with large language models},
  author={Romera-Paredes, Bernardino and Barekatain, Mohammadamin and Novikov, Alexander and Balog, Matej and Kumar, M Pawan and Dupont, Emilien and Ruiz, Francisco JR and Ellenberg, Jordan S and Wang, Pengming and Fawzi, Omar and others},
  journal={Nature},
  volume={625},
  number={7995},
  pages={468--475},
  year={2024},
  publisher={Nature Publishing Group UK London}
}

@inproceedings{liu2024largeLMEA,
  title={Large language models as evolutionary optimizers},
  author={Liu, Shengcai and Chen, Caishun and Qu, Xinghua and Tang, Ke and Ong, Yew-Soon},
  booktitle={2024 IEEE Congress on Evolutionary Computation (CEC)},
  pages={1--8},
  year={2024},
  organization={IEEE}
}

@article{yin2025optimizingllamea,
  title={Optimizing Photonic Structures with Large Language Model Driven Algorithm Discovery},
  author={Yin, Haoran and Kononova, Anna V and B{\"a}ck, Thomas and van Stein, Niki},
  journal={arXiv preprint arXiv:2503.19742},
  year={2025}
}

@article{culberson1998pattern,
  title={Pattern databases},
  author={Culberson, Joseph C and Schaeffer, Jonathan},
  journal={Computational Intelligence},
  volume={14},
  number={3},
  pages={318--334},
  year={1998},
  publisher={Wiley Online Library}
}

@article{ratner1990n2,
  title={The (n2- 1)-puzzle and related relocation problems},
  author={Ratner, Daniel and Warmuth, Manfred},
  journal={Journal of Symbolic Computation},
  volume={10},
  number={2},
  pages={111--137},
  year={1990},
  publisher={Elsevier}
}

@article{felner2004additive,
  title={Additive pattern database heuristics},
  author={Felner, Ariel and Korf, Richard E and Hanan, Sarit},
  journal={Journal of Artificial Intelligence Research},
  volume={22},
  pages={279--318},
  year={2004}
}

@inproceedings{sim2025beyond,
  title={Beyond the Hype: Benchmarking LLM-Evolved Heuristics for Bin Packing},
  author={Sim, Kevin and Renau, Quentin and Hart, Emma},
  booktitle={International Conference on the Applications of Evolutionary Computation (Part of EvoStar)},
  pages={386--402},
  year={2025},
  organization={Springer}
}

@article{min2023beyond,
  title={Beyond accuracy: Evaluating self-consistency of code large language models with identitychain},
  author={Min, Marcus J and Ding, Yangruibo and Buratti, Luca and Pujar, Saurabh and Kaiser, Gail and Jana, Suman and Ray, Baishakhi},
  journal={arXiv preprint arXiv:2310.14053}    ,
  year={2023}
}

@inproceedings{bomer2024sorting,
  title={Sorting Multibay Block Stacking Storage Systems with Multiple Robots},
  author={B{\"o}mer, Thomas and Koltermann, Nico and Pfrommer, Jakob and Meyer, Anne},
  booktitle={International Conference on Computational Logistics},
  pages={34--48},
  year={2024},
  organization={Springer},
    url = {https://doi.org/10.1007/978-3-031-71993-6_3}         
}

@article{wang2024policy,
  title={A policy-based Monte Carlo tree search method for container pre-marshalling},
  author={Wang, Ziliang and Zhou, Chenhao and Che, Ada and Gao, Jingkun},
  journal={International Journal of Production Research},
  volume={62},
  number={13},
  pages={4776--4792},
  year={2024},
  publisher={Taylor \& Francis}
}

@article{hottung2020deep,
  title={Deep learning assisted heuristic tree search for the container pre-marshalling problem},
  author={Hottung, Andr{\'e} and Tanaka, Shunji and Tierney, Kevin},
  journal={Computers \& Operations Research},
  volume={113},
  pages={104781},
  year={2020},
  publisher={Elsevier}
}

@article{burke2013hyper,
  title={Hyper-heuristics: A survey of the state of the art},
  author={Burke, Edmund K and Gendreau, Michel and Hyde, Matthew and Kendall, Graham and Ochoa, Gabriela and {\"O}zcan, Ender and Qu, Rong},
  journal={Journal of the Operational Research Society},
  volume={64},
  number={12},
  pages={1695--1724},
  year={2013},
  publisher={Taylor \& Francis}
}

@article{dat2024hsevo,
  title={HSEvo: Elevating Automatic Heuristic Design with Diversity-Driven Harmony Search and Genetic Algorithm Using LLMs},
  author={Dat, Pham Vu Tuan and Doan, Long and Binh, Huynh Thi Thanh},
  journal={arXiv preprint arXiv:2412.14995}    ,
  year={2024}
}
